\documentclass[twocolumn, switch]{article} %

\usepackage{preprint}

\usepackage{amsmath, amsthm, amssymb, amsfonts}

\usepackage[numbers,square]{natbib}
\usepackage[utf8]{inputenc}	%
\usepackage[T1]{fontenc}	%
\usepackage{xcolor}		%
\usepackage[colorlinks = true,
            linkcolor = purple,
            urlcolor  = blue,
            citecolor = cyan,
            anchorcolor = black]{hyperref}	%
\usepackage{booktabs} 		%
\usepackage{nicefrac}		%
\usepackage{microtype}		%
\usepackage{lineno}		%
\usepackage{float}			%

\usepackage{lipsum}		%

\usepackage{newfloat}
\DeclareFloatingEnvironment[name={Supplementary Figure}]{suppfigure}
\usepackage{sidecap}
\sidecaptionvpos{figure}{c}

\usepackage{titlesec}
\titlespacing\section{0pt}{12pt plus 3pt minus 3pt}{1pt plus 1pt minus 1pt}
\titlespacing\subsection{0pt}{10pt plus 3pt minus 3pt}{1pt plus 1pt minus 1pt}
\titlespacing\subsubsection{0pt}{8pt plus 3pt minus 3pt}{1pt plus 1pt minus 1pt}

\usepackage{times}
\usepackage{helvet}
\usepackage{courier}

\usepackage{amsmath}
\usepackage{txfonts}
\usepackage{amsfonts}
\usepackage{algorithm,algpseudocode}%
\usepackage{multirow}
\usepackage{paralist}
\usepackage[subrefformat=parens]{subcaption}
\usepackage{bm}
\usepackage{lscape}

\newcommand{\Zu}[1]{Figure \ref{fig:#1}}

\newcommand{\Shiki}[1]{Eq \eqref{eq:#1}}

\newcommand{\Algo}[1]{Algorithm \ref{alg:#1}}

\newcommand{\argmax}{\mathop{\rm arg~max}\limits}
\newcommand{\argmin}{\mathop{\rm arg~min}\limits}

\algnewcommand{\Inputs}[1]{%
  \State \textbf{Inputs:}
  \Statex \hspace*{\algorithmicindent}\parbox[t]{.8\linewidth}{\raggedright #1}
}

\algnewcommand{\Outputs}[1]{%
  \State \textbf{Outputs:}
  \Statex \hspace*{\algorithmicindent}\parbox[t]{.8\linewidth}{\raggedright #1}
}

\algnewcommand{\Initialize}[1]{%
  \State \textbf{Initialize:}
  \Statex \hspace*{\algorithmicindent}\parbox[t]{.8\linewidth}{\raggedright #1}
}

\algnewcommand{\Estimate}[1]{%
  \State \textbf{Estimate:}
  \Statex \hspace*{\algorithmicindent}\parbox[t]{.8\linewidth}{\raggedright #1}
}
\title{Iterative Machine Teaching without Teachers}

\usepackage{authblk}

\author[1\thanks{\tt{mingzhe.yang@hcomp.cs.tsukuba.ac.jp}}]{Mingzhe Yang}
\author[1]{Yukino Baba}
\affil[1]{University of Tsukuba}

\begin{document}

\twocolumn[ %
  \begin{@twocolumnfalse} %
\maketitle

\begin{abstract}
  Iterative machine teaching is a method for selecting an optimal teaching example that enables a student to efficiently learn a target concept at each iteration. Existing studies on iterative machine teaching are based on supervised machine learning and assume that there are teachers who know the true answers of all teaching examples. In this study, we consider an unsupervised case where such teachers do not exist; that is, we cannot access the true answer of any teaching example. Students are given a teaching example at each iteration, but there is no guarantee if the corresponding label is correct. Recent studies on crowdsourcing have developed methods for estimating the true answers from crowdsourcing responses. In this study, we apply these to iterative machine teaching for estimating the true labels of teaching examples along with student models that are used for teaching. Our method supports the collaborative learning of students without teachers. The experimental results show that the teaching performance of our method is particularly effective for low-level students in particular. 

\end{abstract}
\vspace{0.35cm}

  \end{@twocolumnfalse} %
] %

\noindent 
\section{Introduction}

Iterative machine teaching~\cite{liu2017iterative} is a method for selecting an optimal teaching example that enables a student to efficiently learn a target concept.
A teacher in iterative machine teaching can access both the true classification model and the student models. 
At each iteration, the teacher selects a teaching sample that will make the student model closest to the true model. 
The student model is then updated according to the selected sample. 
Existing studies on iterative machine teaching are based on supervised machine learning, in which a student model corresponds to a model trained in machine learning. 
As with supervised machine learning, iterative machine teaching assumes that they are given a teaching set, which consists of a pair of a sample and a true label (i.e., $(\bm{x}, y)$). 
The teaching set is used for teaching students as well as for estimating the true model.

In this study, we consider a case where the true label is not given for any example in the teaching set and investigate whether iterative machine teaching works without true labels. 
This situation can be occur when students collaboratively teach each other; students learn by using the ways in which others answer as teaching examples.
A typical case is crowdsourcing. 
For example, in the Galaxy Zoo project~\cite{willett2013galaxy}, novice workers learned how to classify galaxy images using the answers of experienced workers.

Our approach for estimating the true labels is to use student answers. 
There have been several attempts in crowdsourcing research to estimate the true answers for a question using the answers from crowdsourcing workers~\cite{zheng2017truth}. 
We especially focus on the ``learning from crowds'' methods~\cite{raykar2010learning,kajino2012convex}, which estimate a true classification model from crowdsourcing labels as ordinary machine learning methods estimate the model from the true labels.
In addition, these methods estimate the classification model of each worker (student); we use this model to choose a suitable teaching example for each student. 
In our iterative machine teaching procedure, we occasionally ask students to take an exam. 
The answers are then used to estimate both the true model and the student models. 
At each iteration, we select a teaching example for each student according to the estimated true model and the student model.
The label of the teaching example is inferred using the estimated true model.

We conducted experiments with two datasets and found that the proposed method achieves a high level of learning efficiency. 
We also show that the proposed method allows students to learn efficiently even when the number of answers from students is small.

The contributions of our work are summarized as follows:
\begin{itemize}
    \item We identify a novel problem setting for iterative machine teaching without the true labels.
    \item We propose an iterative machine teaching procedure that estimates the true labels by using student answers.
    \item We show that the proposed method improves learning efficiency especially for low-level students.
\end{itemize}

\section{Related Work}
Machine teaching is an inverse problem of machine learning where machines teach humans instead of teaching machines.
Several methods for choosing a set of teaching examples have been studied. 
For example, Singla et al. proposed a noise-tolerant model of the learning process in classification tasks~\cite{singla2014near}. 
Another study examined a framework for finding a suitable example set that trains students effectively~\cite{zhu2015machine}. 

Liu et al. formulated an iterative framework in which a teacher selects a single teaching example at each iteration~\cite{liu2017iterative}, enabling students to learn faster than before. 
This method assumes that a teacher is able to access the student models. 
Moreover, Liu et al. proposed the treatment of a student model as a black-box model~\cite{liu2018towards}. 
Instead of the actual student models, a teacher of this algorithm has virtual student models and selects an example for each virtual student. 
The idea of having virtual student models is similar to our method; however, this method still depends on a labeled teaching set, whereas our method uses an unlabeled teaching set. 

Zhou et al. considered the extent to which students remembered the previous examples and incorporated the forgetting rate into an iterative teaching algorithm~\cite{zhou2018unlearn}.
The authors then used an algorithm~\cite{zhu2003semi} to estimate the true model. 
The teaching examples were chosen according to the forgetting rates and the true model.
Similar to the proposed method, this method estimated the true model. However, it did not estimate the student models and used a labeled teaching set.

Curriculum learning is a general training strategy that presents teaching examples ranging from easy to difficult~\cite{bengio2009curriculum}. It is based on the natural flow of human learning, which begins with the simplest examples and tackles the more difficult examples over time.

\section{Preliminary}
\subsection{Iterative Machine Teaching}
    We first introduce the problem setting for iterative machine teaching~\cite{liu2017iterative}. We denote $\mathcal{X} \subset \mathbb{R}^d$ as the $d$-dimensional feature representations of all instances and $\mathcal{Y}$ as the set of labels. The teacher can access a labeled subset $\Phi \subset \mathcal{X} \times \mathcal{Y}$, which is called a teaching set. 
    We specifically focus on binary concept learning; $\bm{x} \in \mathcal{X}$ is the feature of an example and $y \in \{0, 1\}$ is its corresponding binary class label. 
    
    At each iteration, the teacher selects one example $(\bm{x}, y) \in \Phi$ for a student. 
    The goal of iterative machine teaching is to select an example in which the student efficiently learns the target concept.

    \subsection{Omniscient Teacher Algorithm}
    The omniscient teacher algorithm~\cite{liu2017iterative} is an algorithm for iterative machine teaching. 
    This algorithm assumes that the teacher has access to the true classification model, $\bm{w}^\star$, and the initial classification model of each student $j$, $\bm{w}^0_j$. 
    In practical terms, we can estimate $\bm{w}^\star$ by using the teaching set, and estimate $\bm{w}^0_j$ by obtaining initial answers for the teaching set from each student.
    
    The omniscient teacher algorithm assumes that the students update their model according to stochastic gradient descent (SGD). That is, after the student $j$ is presented an example $(\bm{x}, y) \in \Phi$ at the $t$-th iteration, the student will update the model from $\bm{w}^t_j$ to $\bm{w}^{t+1}_j$ as follows:
    \begin{equation}
        \bm{w}^{t+1}_j=\bm{w}^{t}_j-\alpha_{t} \frac{\partial \mathcal{L}(\langle \bm{w}_j^t, \bm{x}\rangle, y)}{\partial \bm{w}_j^t} 
        \label{eq:sgd}
    \end{equation}
    where $\alpha_t$ is the learning rate and $\mathcal{L}(\bm{w}, \bm{x})$ is the loss function. 
    Because we focus on binary classification, we employ the logistic loss.

    The best example will most correctly update the student model. 
    In other words, if the model is updated according to the best example, the distance between the updated model, $\bm{w}^{t+1}_j$, and the true model, $\bm{w}^\star$, will be minimized.
    Using \Shiki{sgd}, the Euclidean distance between them is given as follows:
    \begin{align}
        &\left\|\bm{w}^{t+1}_j-\bm{w}^{\star}\right\|_{2}^{2}\nonumber \\
        =&\left\|\bm{w}^{t}_j-\alpha_{t} \frac{\partial \mathcal{L}(\langle \bm{w}^t_j, \bm{x}\rangle, y)}{\partial \bm{w}^t_j}-\bm{w}^{\star}\right\|_{2}^{2} \nonumber \\
        =&\left\|\bm{w}^{t}_j-\bm{w}^{\star}\right\|_{2}^{2}+\alpha_{t}^{2} \quad\left\|\frac{\partial \mathcal{L}\left(\left\langle \bm{w}^{t}_j, \bm{x}\right\rangle, y\right)}{\partial \bm{w}^{t}_j}\right\|_{2}^{2} \nonumber \\
        \hspace{5mm} & -2 \alpha_{t}\left\langle \bm{w}^{t}_j-\bm{w}^{\star}, \frac{\partial \mathcal{L}\left(\left\langle \bm{w}^{t}_j, \bm{x}\right\rangle, y\right)}{\partial \bm{w}^{t}_j}\right\rangle.
        \label{eq:IMT}
    \end{align}
    
    The omniscient teacher algorithm selects the example by solving the following optimization problem:
    \footnotesize
    \begin{align}
        \argmin_{(\bm{x},y) \in \Phi} \alpha_{t}^{2}\quad\left\|\frac{\partial \mathcal{L}\left(\left\langle \bm{w}^{t}_j, \bm{x}\right\rangle, y\right)}{\partial \bm{w}^{t}_j}\right\|_{2}^{2} -2 \alpha_{t}\left\langle \bm{w}^{t}_j-\bm{w}^{\star}, \frac{\partial \mathcal{L}\left(\left\langle \bm{w}^{t}_j, \bm{x}\right\rangle, y\right)}{\partial \bm{w}^{t}_j}\right\rangle. \label{eq:omt}
    \end{align}
    \normalsize
    
    Note that $\left\|\bm{w}^{t}_j-\bm{w}^{\star}\right\|_{2}^{2}$ in \Shiki{IMT} is omitted from the optimization because this term is not related to $(\bm{x}, y)$.

\section{Iterative Machine Teaching with Unlabeled Teaching Set}
\subsection{Problem Setting}
The existing algorithms for iterative machine teaching assume that they are given a labeled teaching set.
We consider a different problem setting in which we are given an \textit{unlabeled} teaching set, that is, our teaching set is simply a subset of $\mathcal{X}$ and the true label for each $\bm{x}$ is not given. 
We denote such a teaching set as $\Phi_x \subset \mathcal{X}$.
In this setting, we are not able to access the true model and the student models, which are estimated by using the labeled teaching set in the existing problem setting. 

\Zu{imt} illustrates the process of iterative machine teaching with the labeled teaching set. 
Here, we can consider that there is an oracle that provides the information about the true model and the student models; the teacher uses this information to select a teaching example. 

Our idea for addressing iterative machine teaching with the unlabeled teaching set is to ask students to take an exam at each iteration and to use their answers to estimate both the true model and the student models. 
In other words, we use the answers from the students instead of the information given by the oracle, as illustrated in \Zu{wot}. 

Formally, we have an exam set, $X \subset \mathcal{X}$, and the students are asked to answer a label for each $\bm{x}_i \in X$ at each exam. 
The label from the student $j$ to the instance $x_i$ is denoted by $y_{ij} \in \{0, 1\}$, and all the labels are denoted by $Y = \{y_{ij}\}_{i,j}$. 
At each iteration, given a teaching set $\Phi_x \subset \mathcal{X}$, an exam set $X \subset \mathcal{X}$, and the answers $Y$, our goal is to choose a teaching example $\bm{x} \in \Phi_x$ for a student. 

\begin{figure*}[tbp]
    \begin{minipage}{0.49\hsize}
        \centering
        \includegraphics[width = 6cm]{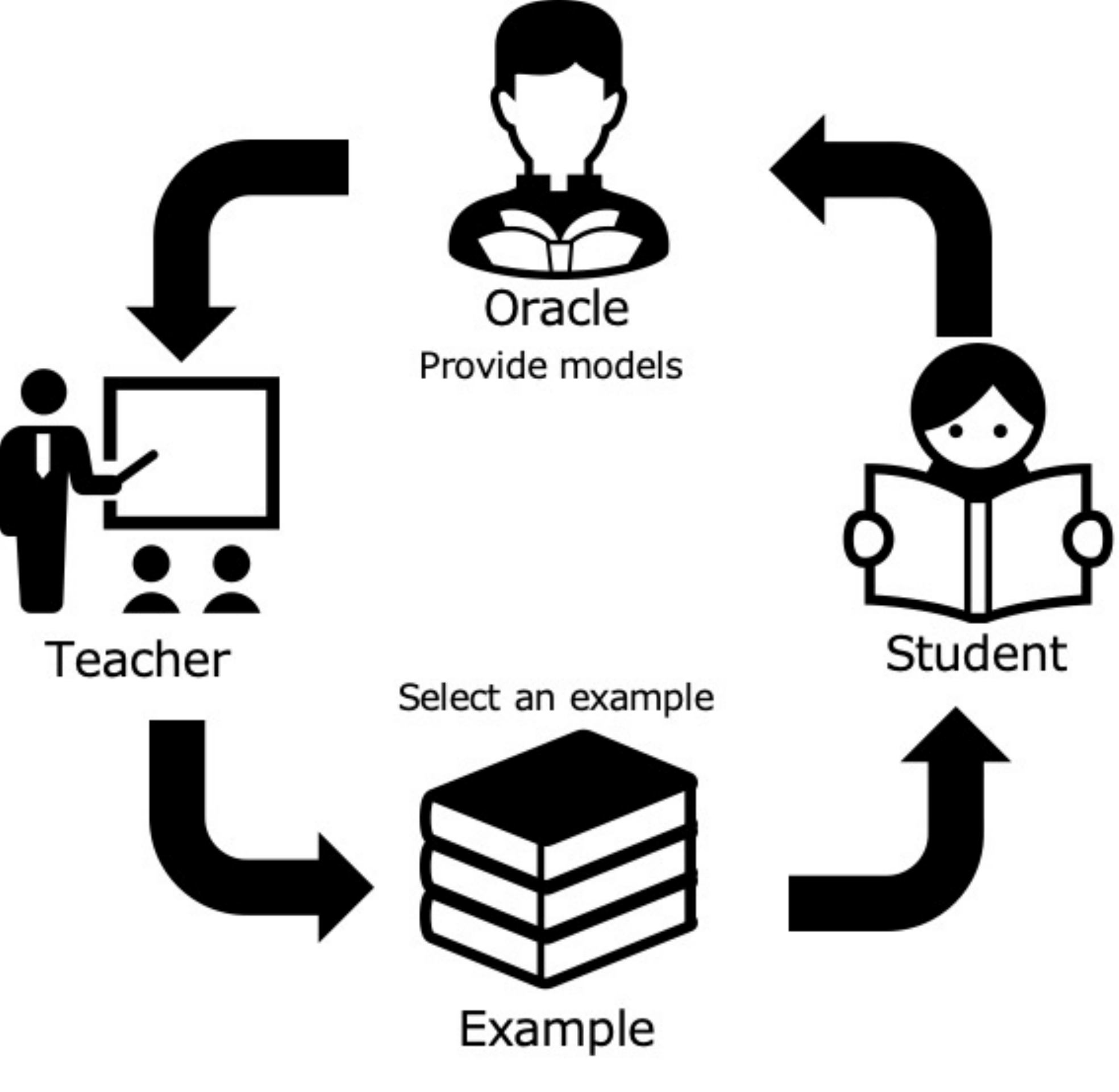}
        \caption{Procedure of Omniscient teacher algorithm}
        \label{fig:imt}
    \end{minipage}
    \begin{minipage}{0.49\hsize}
        \centering
        \includegraphics[width = 6cm]{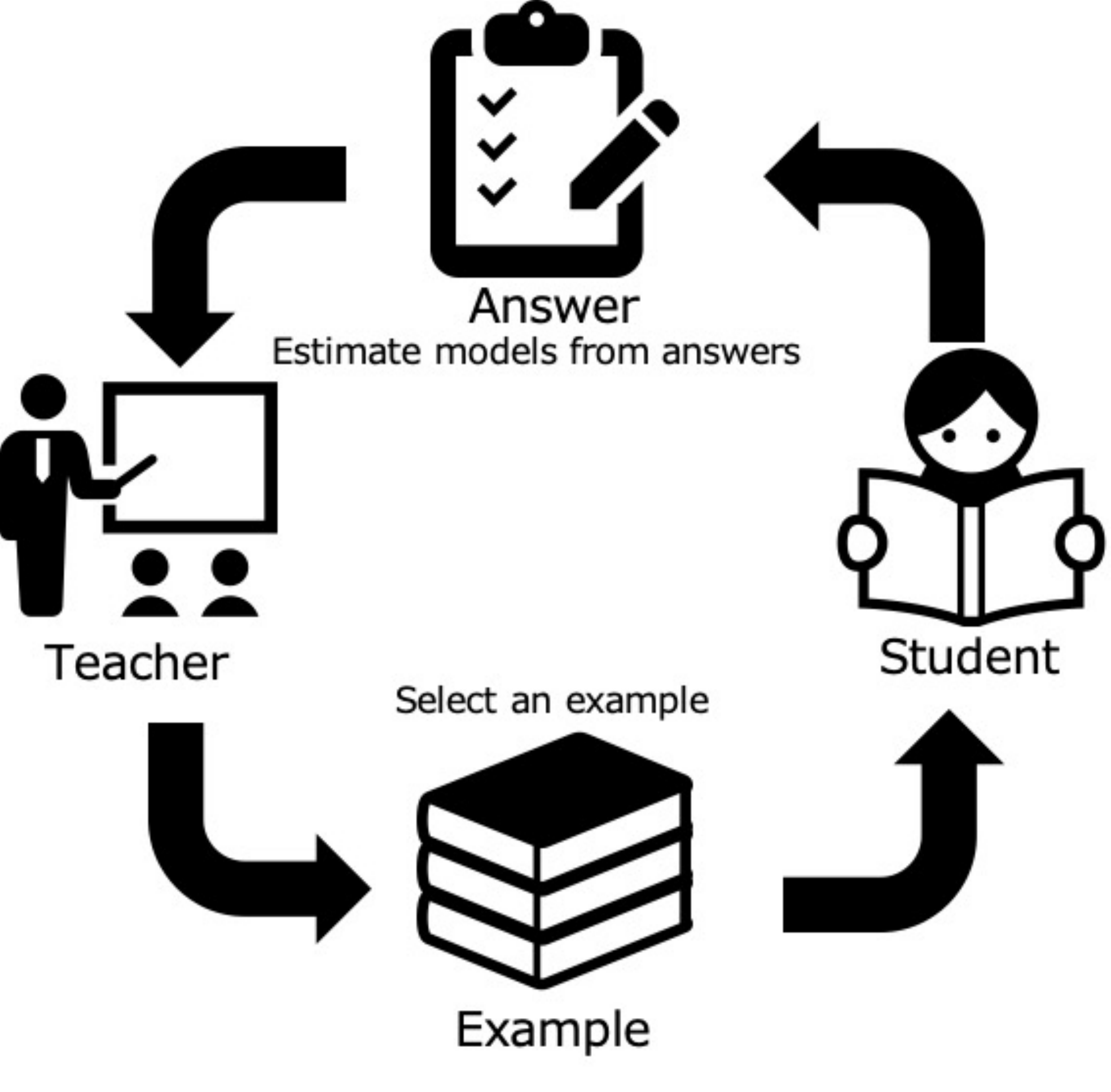}
        \caption{Procedure of proposed algorithm}
        \label{fig:wot}
    \end{minipage}
\end{figure*}

\subsection{Model Estimation}
Learning from crowds~\cite{raykar2010learning,kajino2012convex} approaches are methods for estimating the true model and the worker models by using the worker responses. 
We apply one of these methods~\cite{kajino2012convex} for iterative machine teaching. 
Based on this method, we assume that the true model is generated from a Gaussian distribution:
\begin{align}
    \operatorname{Pr}\left[\bm{w}^{\star} | \eta\right]&=\mathcal{N}\left(\mathbf{0}, \eta^{-1} \mathbf{I}\right)\label{eq:w_star},
\end{align}
where $\eta > 0$ is a hyperparameter.

We also assume that each student model at an iteration, $\bm{w}_j$, is generated by adding a Gaussian noise to the true model $\bm{w}^\star$:
\begin{align}
    \operatorname{Pr}\left[\bm{w}_{j} | \bm{w}^{\star}, \lambda\right]&=\mathcal{N}\left(\bm{w}_{j} \mid \bm{w}^{\star}, \lambda^{-1} \mathbf{I}\right)\label{eq:wj},
\end{align}
where $\lambda >0$ is a hyperparameter. 
Finally, the answer $y_{ij}$ is generated according to the student model: 
\begin{align}
    \operatorname{Pr}\left[y_{j}=1 | \bm{x}_i, \bm{w}_{j}\right]=\sigma\left(\langle \bm{w}_j,\bm{x}_i\rangle\right),
\end{align}
where $\sigma(x) = \left(1+\exp(-x)\right)^{-1}$ is the sigmoid function.

By denoting $\mathbf{W}=\left\{\mathbf{w}_{j}\right\}_j$, the posterior distribution of $\bm{w}^\star$ and $\bm{W}$, given the observation $X$ and $Y$, is formalized as follows:
\begin{align}
        &\operatorname{Pr}\left[\bm{W}, \bm{w}^{\star} | X, Y, \eta, \lambda\right]\nonumber \\
        &\propto \operatorname{Pr}[Y | \bm{W}, X] \cdot \operatorname{Pr}\left[\bm{W} | \bm{w}^{\star}, \lambda\right] \cdot \operatorname{Pr}\left[\bm{w}^{\star} | \eta\right].
\end{align}

Let $F(\bm{w}^\star,\bm{W})$ be the negative log-posterior distribution of $\bm{w}^\star$ and $\bm{W}$ without constants, which is given as follows:
\footnotesize
\begin{align}
    &F\left(\bm{w}^{\star}, \bm{W}\right)\label{eq:F()}\nonumber\\
    =&\log \operatorname{Pr}[Y | \bm{W}, X] + \log \operatorname{Pr}\left[\bm{W} | \bm{w}^{\star}, \lambda\right] + \log \operatorname{Pr}\left[\bm{w}^{\star} | \eta\right]\nonumber\\
    =&\frac{\lambda}{2} \sum_{j}\left\|\bm{w}_{j}-\bm{w}^{\star}\right\|^{2}+\frac{\eta}{2}\left\|\bm{w}^{\star}\right\|^{2}\nonumber\\
    &-\sum_{i,j}\left[y_{i j} \log \sigma\left(\langle\bm{w}_{j}, \bm{x}_{i}\rangle\right)+\left(1-y_{i j}\right) \log \left(1-\sigma\left(\langle\bm{w}_{j}, \bm{x}_{i}\rangle\right)\right)\right].
\end{align}
\normalsize
The maximum-a-posterior estimators $\hat{\bm{w}}^{\star}$ and $\hat{\bm{W}}$ are obtained by solving an optimization problem: 
\begin{align}
    \hat{\bm{w}}^{\star}, \hat{\bm{W}} = \argmax_{\bm{w}^\star, \bm{W}} F\left(\mathbf{w}^\star, \bm{W}\right).
\end{align}
We obtain the estimated true model $\hat{\bm{w}}^{\star}$ and the student models $\hat{\bm{W}} = \hat{\bm{w}}_j$ after the students take an exam. The algorithm used for model estimation is summarized in \Algo{estimate}.

\begin{algorithm}[tb]
  \caption{Model estimation}
  \label{alg:estimate}
  \begin{algorithmic}[1]
    \Inputs{exam set $X$; student answers $Y$;\\ hyperparameters $\eta$ and $\lambda$}
    \Outputs{estimated true model $\hat{\bm{w}}^\star$;\\ estimated student models $\hat{\bm{W}}$}
    \Function{EstimateModel}{$X,Y,\eta,\lambda$}
    \State$\hat{\bm{w}}^\star,\hat{\bm{W}}=\min _{\hat{\bm{w}}^\star, \hat{\bm{W}}} F\left(\hat{\bm{w}}^\star, \hat{\bm{W}}\right)$
    \State return $\hat{\bm{w}}^\star,\hat{\bm{W}}$
    \EndFunction
  \end{algorithmic}
\end{algorithm}

\subsection{Teaching Procedure}
After estimating the true model and the student models, we can simply apply the omniscient teacher algorithm (Section 3.2) by replacing $\bm{w}^\star$ with $\hat{\bm{w}}^\star$, and $\bm{w}^t_j$ with $\hat{\bm{w}}_j$, respectively. 

Because our teaching set is unlabeled, we must estimate the true label for each example in the teaching set to allows students to learn. 
This is done by simply using the estimated true model $\hat{\bm{w}}^\star$. At each iteration, we estimate the true label for an instance $\bm{x}_i \in \Phi_x$ by calculating $\sigma\left(\langle \bm{w}^\star,\bm{x}_i\rangle\right)$. If this value is larger than $0.5$, the estimated true label is $\hat{y}_i = 1$; otherwise $\hat{y}_i = 0$.
The overall procedure is summarized in \Algo{without teacher}.

\begin{algorithm}[tb]
  \caption{Label estimation}
  \label{alg:get label}
  \begin{algorithmic}
    \Inputs{estimated true model $\hat{\bm{w}}^\star$;  teaching instances $\Phi_x$}
    \Outputs{Pseudo teaching set $\hat{\Phi}$}
    \Function{GetPseudoTeachingSet}{$\hat{\bm{w}}^\star$, $\Phi_x$}
    \State $\hat{\Phi} = \emptyset$
    \For{each instance $\bm{x}_i \in \Phi$}
    \State $\hat{y}_i= \left\{ \begin{array}{ll}
        1 & \sigma{(\langle \hat{\bm{w}}^\star,\bm{x_i}\rangle)}> 0.5 \\
        0 & (otherwise)
    \end{array} \right.$
    \State add $(\bm{x}_i, \hat{y}_i)$ to $\hat{\Phi}$
    \EndFor
    \State return $\hat{\Phi}$
    \EndFunction
  \end{algorithmic}
\end{algorithm}

\begin{algorithm}[tb]
  \caption{Iterative Machine Teaching without Teachers}
  \label{alg:without teacher}
  \begin{algorithmic}[1]
    \Inputs{Teaching set $\Phi_x$; exam set $X$;\\
    hyper parameters $\alpha_t$, $\eta$ and $\lambda$; max number of iterations $\textsc{MaxIter}$; number of exam samples $T$.}
    \Initialize{$t=1$\\
                $\hat{\bm{w}}^\star \sim \mathcal{N}(\hat{\bm{w}}^\star\mid 0,\lambda^{-1}\bm{I})$}
    \While{$t < \textsc{MaxIter}$}
    \State ask students to take an exam and get answers $Y^t$ for $T$ randomly chosen samples from the exam set $X$.
    \State $\hat{\bm{w}}^{\star t},\hat{\bm{W}}^t=\textsc{EstimateModel}(X, Y^t,\eta,\lambda)$
    \State $\hat{\Phi}^t= \textsc{GetPseudoTeachingSet}(\hat{\bm{w}}^{\star t},\Phi_x)$
    \For{each student $j$}
    \begin{align*}
        (\bm{x}_j^t,\hat{y}_j^t) =& \argmin_{(\bm{x},\hat{y}) \in \hat{\Phi}^t} \alpha_{t}^{2}\quad\left\|\frac{\partial \mathcal{L}\left(\left\langle \hat{\bm{w}}^{t}_j, \bm{x}\right\rangle, \hat{y}\right)}{\partial \hat{\bm{w}}^{t}_j}\right\|_{2}^{2}\\
        &-2 \alpha_{t}\left\langle \hat{\bm{w}}^{t}_j-\hat{\bm{w}}^{\star t}, \frac{\partial \mathcal{L}\left(\left\langle \hat{\bm{w}}^{t}_j, \bm{x}\right\rangle, \hat{y}\right)}{\partial \hat{\bm{w}}^{t}_j}\right\rangle
    \end{align*}
    \State{Show an example $(\bm{x}^t_j, \hat{y}^t_j)$ to the student.}
    \EndFor
    \State $t \leftarrow t + 1$
    \EndWhile
  \end{algorithmic}
\end{algorithm}

\section{Experiments}
\subsection{Datasets}
Here, we demonstrate the effectiveness of the proposed method; the experiments were designed to investigate if students can learn efficiently with the proposed method.

    To verify them, we used two datasets: a synthetic dataset and a wine dataset.
    
    \subsubsection{Insect}
    We designed a synthetic experiments by referring to the experiment conducted by Single et al.~\cite{singla2014near}.
    We generated simplified images of two insect species, the weevil and the Vespula; examples are shown in \Zu{weebilandvespula}. Weevils have heads that are smaller and lighter in color than their bodies, and Vespulas species have larger, dark-colored heads. Weevils and Vespulas are different in two factors: \begin{inparaenum}[(i)]
        \item the head/body size ratio $f_1$
        \item the head/body color ratio $f_2$.
    \end{inparaenum}
    Thus, each image is characterized by the feature vector $\bm{x} = \left[f_1, f_2\right]^\top$.
    We sampled features from $2$D Gaussian distributions with different means $\bm{\mu}$ for both the weevil and Vespula; the mean for weevils was $[-0.10, -0.13]^\top$ and the mean for Vespula was $[0.10, 0.13]^\top$ with the covariance of $0.12\bm{I}$. 
    The sampled features are plotted in \Zu{wvsvandmodel}. 
    We generated $2{,}000$ images in total; 
    $1{,}000$ were weevil images and $1{,}000$ were Vespula images.
    We used $75\%$ of these examples as the teaching set, and the remainder was used for evaluation. 
    
    \subsubsection{Wine}
    We used Wine Quality dataset provided by the UCI Machine Learning Repository~\cite{cortez2009modeling}. 
    Specifically, we used the red wine data, which had $1{,}600$ examples with 11-dimensional feature vectors. 
    Each wine has a quality score between zero and ten as the target value, and we binarized these scores with a threshold of $5$. 
    As mentioned above, we used $75\%$ examples as the teaching set, and the remainder was used for evaluation.
    
    \begin{figure}[tb]
        \centering
        \includegraphics[width = 7cm]{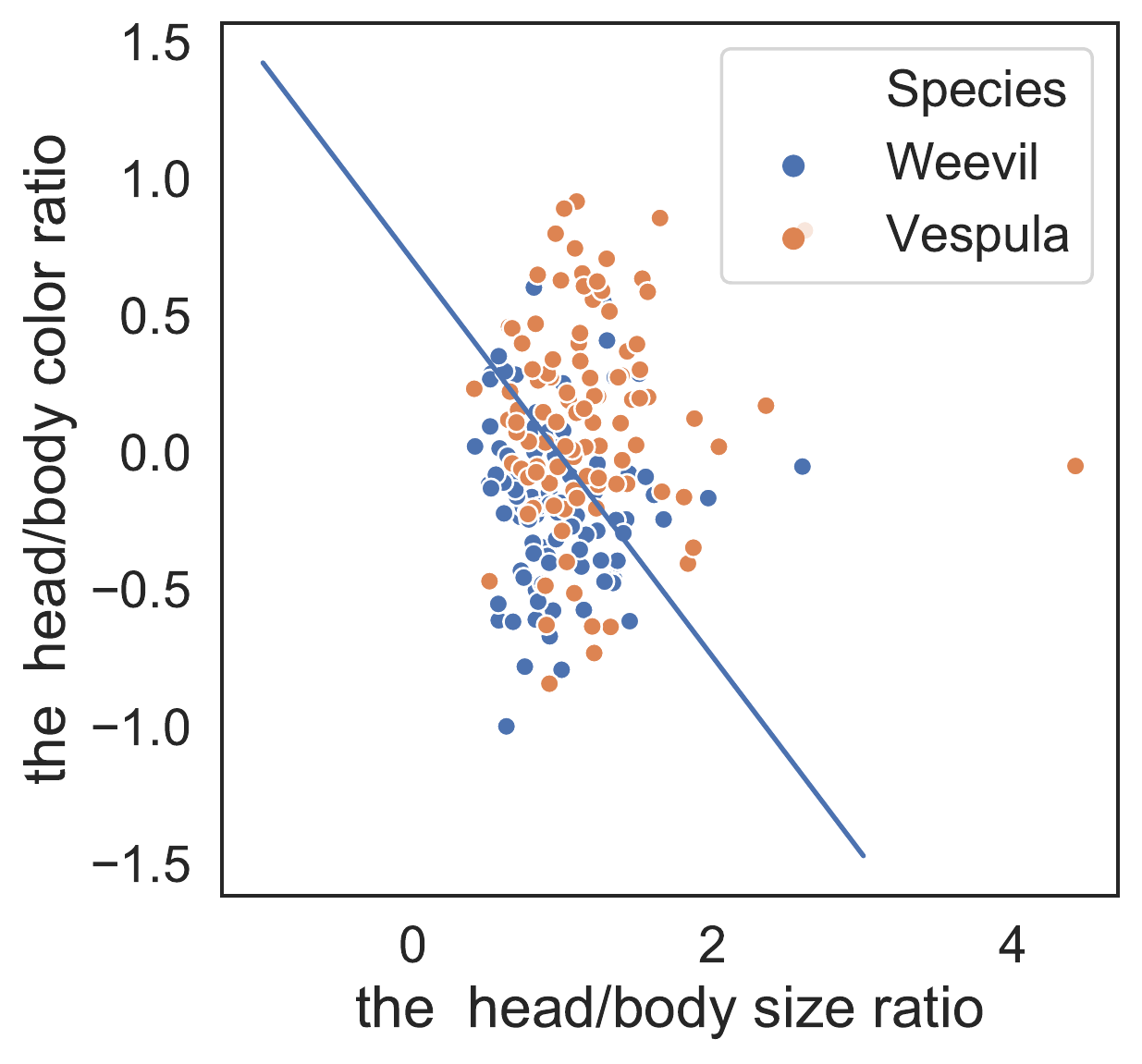}
        \caption{Plot of weevil and Vespula features. The blue line indicates the true classification model ($\bm{w}^\star$).}
        \label{fig:wvsvandmodel}
    \end{figure}
    
    \begin{figure}[tb]
        \centering
        \includegraphics[width = 7cm]{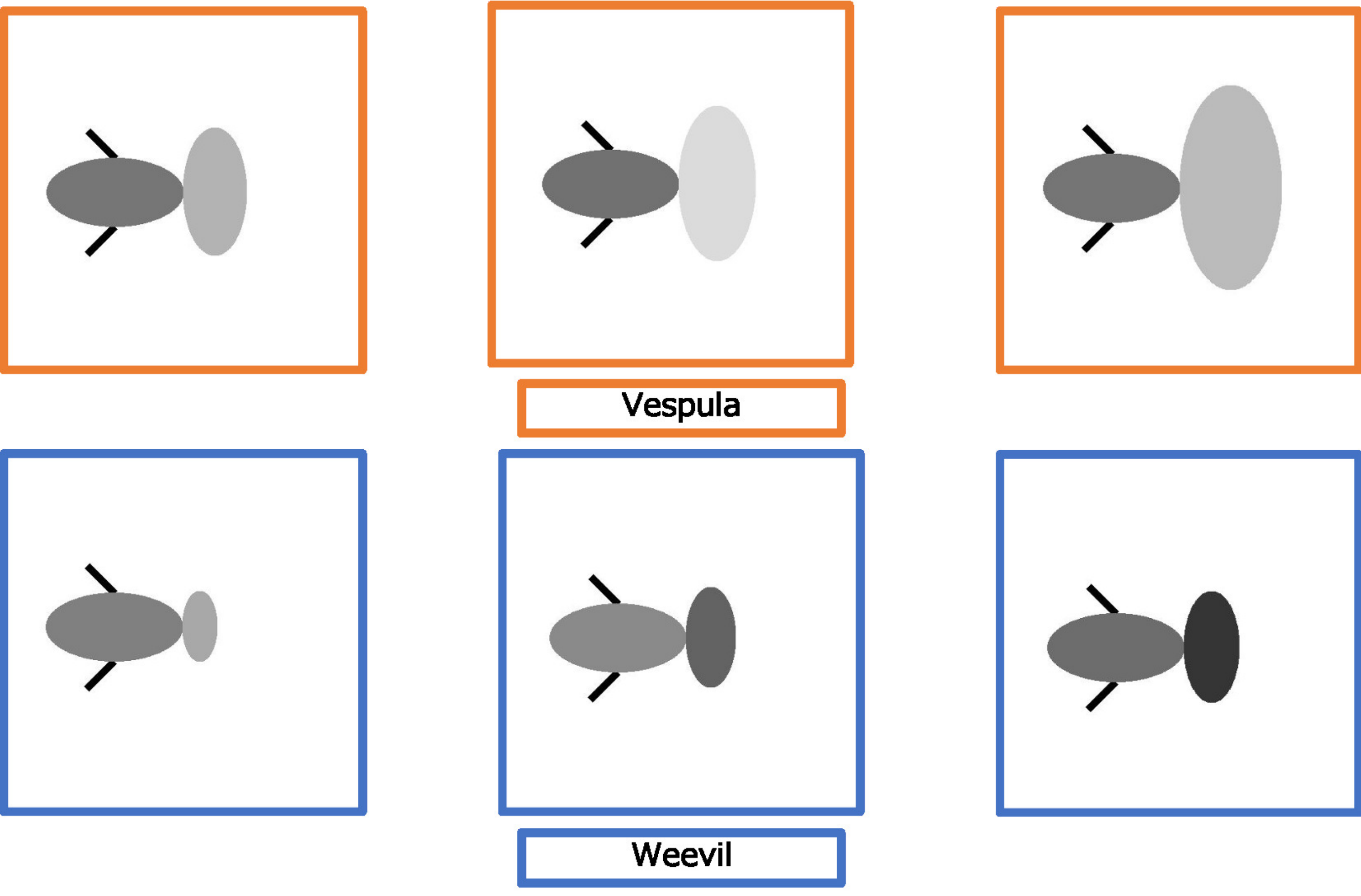}
        \caption{Synthetic images of weevils (top) and Vespulas (bottom)}
        \label{fig:weebilandvespula}
    \end{figure}
    
    \subsection{Student Models}
    We prepared synthetic students for this experiment. We first obtained the true model  $\bm{w}^{\star}=\argmin_{\bm{w}} \sum_i \mathcal{L}(\langle \bm{w}, \bm{x}_i\rangle, y_i)$ by using all the examples. The ROC-AUC score of $\bm{w}^{\star}$ is $0.762$ for the insect dataset and $0.821$ for the wine dataset. 
    We then generated the student models $\bm{W}$ by using \Shiki{wj} with $\lambda \in \{1, 2, 3, 4, 5\}$; we prepared ten groups of students for each $\lambda$ and each group had ten students. 
    Their initial abilities (ROC-AUC scores of $\bm{w}^0_j$) were shown in \Zu{insect_and_wine_box}. With smaller $\lambda$, students are likely to have higher abilities and the variance of them are small.

    We set that students updated their models according to \Shiki{sgd} with $\alpha_t = 0.01$ when they are given a teaching example. 
    We set that students to provide an answer based on $\sigma\left(\langle \bm{w}^t_j,\bm{x}\rangle\right)$ for an instance $\bm{x}$ in the exam set according to their current model $\bm{w}^t_j$. 
    Note that the proposed method does not have access to these parameters; they just used only for simulating students in the experiments. 
    
    \begin{figure}[tb]
        \begin{minipage}{1\hsize}
            \centering
            \includegraphics[width=5cm]{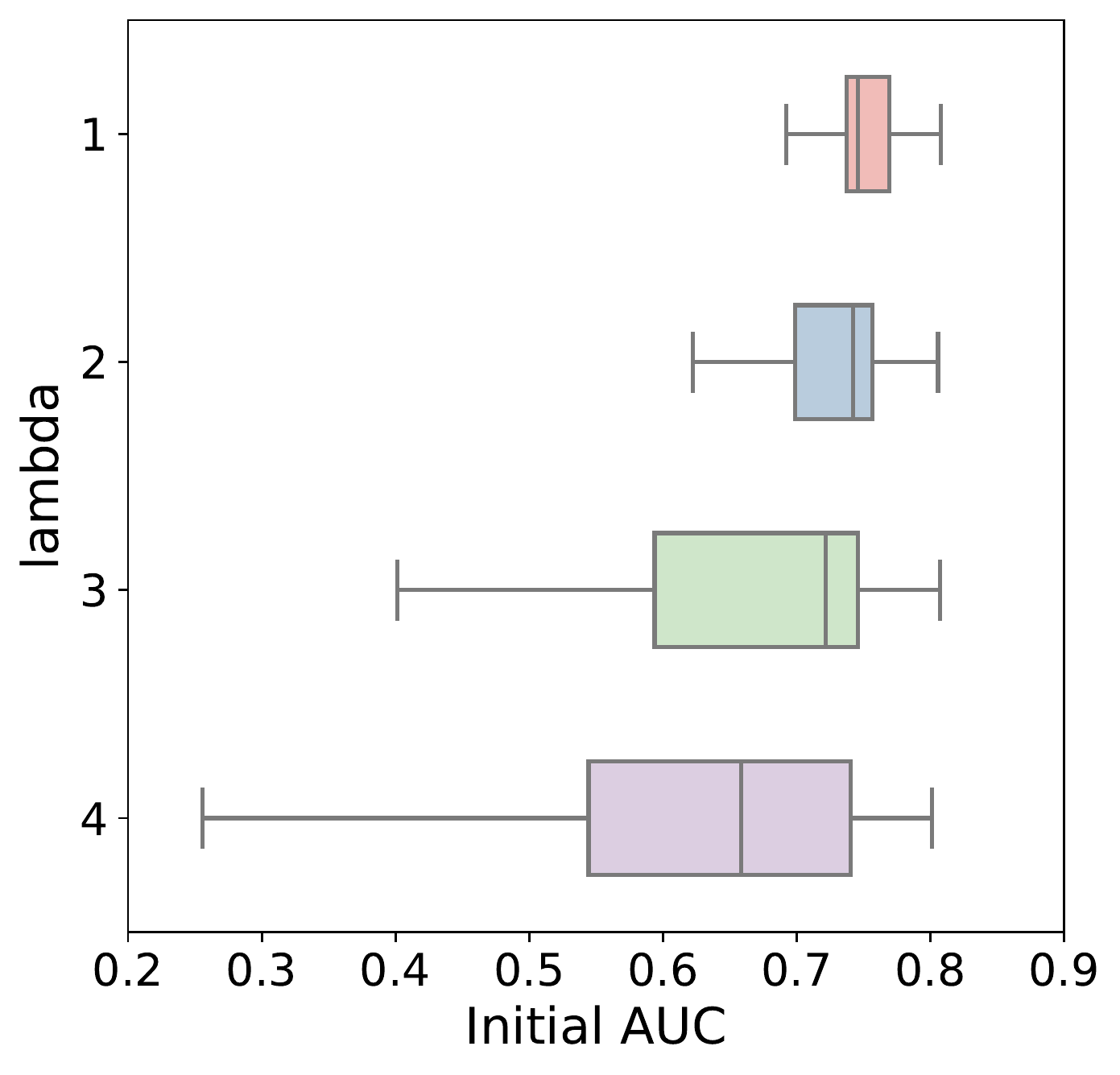}
            \subcaption{insect}
            \label{fig:insect_box}
        \end{minipage}
        \begin{minipage}{1\hsize}
            \centering
            \includegraphics[width=5cm]{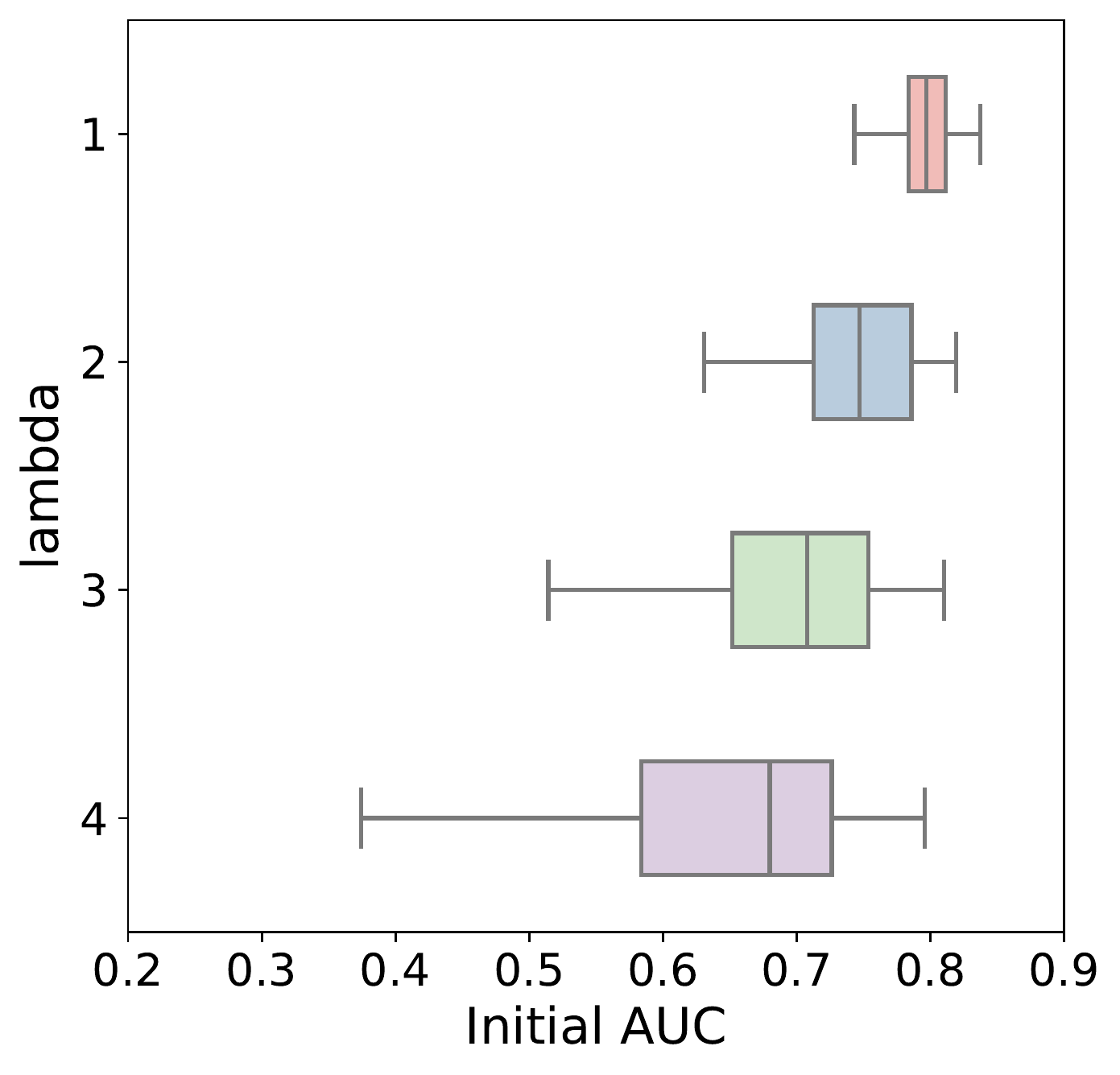}
            \subcaption{wine}
            \label{fig:wine_box}
        \end{minipage}
          \caption{Initial student abilities according to $\lambda$}
        \label{fig:insect_and_wine_box}
    \end{figure}

    \subsection{Baselines and Setup}
    We compared the proposed method with two baselines:
    \begin{itemize}
        \item \textbf{Random teacher}: This algorithm randomly selects a teaching example at each iteration.
        \item \textbf{Omniscient teacher}~\cite{liu2017iterative}: This algorithm selects a teaching example by using the true model and the student models, while the proposed method estimates these models.
        In other words, the omniscient teacher algorithm can access to the models described in Section 5.2.
        This algorithm also has a labeled teaching set; thus, the students learned using this algorithm are presented the true label.
    \end{itemize}
    All the methods selected one teaching example at each iteration for each student. 
    Considering the practical situation, we showed the same teaching example to each student once.
    
    The proposed method and the omniscient teacher algorithm both set the learning rate $\alpha_t$ to $0.01$; the hyperparameter of the proposed method was set to $\eta=1$;
    the number of answers at each iteration was set to $T=100$.

    \subsection{Results}
    \subsubsection{Learning efficiency.}
    We compared the learning efficiency of students with different teaching algorithms. 
    We applied the algorithms to each student group and simulated their learning process using these algorithms.
    For each group, we randomly chose examples for the teaching set, that were also used as the exam set for obtaining student answers.

    \Zu{insect_all} and \Zu{wine_all} show the average ROC-AUC scores of the student models $\{\bm{w}_j^t\}$ at each iteration $t$. 
    The ROC-AUC scores were calculated based on the evaluation set, which was never shown to the students. 
    Our proposed method could select samples that had a measurable learning effect even though it does not use the true labels. 
    However, when $\lambda$ was large, our method selected them as effectively as the random teacher. 
    Our algorithm showed better performance with smaller $\lambda$ values; it is because our method requires a few high ability students in a group for estimating true labels.
    
    \subsubsection{Influence of student initial abilities.}
    We separated students into three categories according to their initial abilities; The top $25\%$ students were ``high-level'', and the bottom $25\%$ students were ``low-level''; the others were considered as ``middle-level''.
   \Zu{insect_high}, \Zu{insect_middle}, and \Zu{insect_low} shows the ROC-AUC scores of high-, middle-, and low-level students for the insect dataset, respectively. 
   In the insect dataset, all the methods were not effective for high-level students. 
    Their initial abilities were already high and they abilities did not improve even though when we used the omniscient teacher. 
    We find that the proposed method was particularly effective for low-level students when $\lambda$ was small;
    that is, our method supported them to learn by the answers of high-ability students. 
    
    \Zu{wine_high}, \Zu{wine_middle}, and \Zu{wine_low} are the results for the wine dataset. 
    In the wine dataset, when $\lambda=1$, the performance of high-ability students became worse by using the proposed method.
    When some true labels are estimated wrongly, student models are easily updated in a wrong direction. 
    This has a severe effect on the learning of high-ability students.

\subsubsection{Influence of exam interval.}

The proposed method asked students to take an exam at each iteration to estimate their models at the time. 
In order to make this process more efficient, we examined the learning effect with with different test interval $B$; the students took an exam once every $B$ iterations.
\Zu{insect_B} shows the average ROC-AUC score with different $B$.
We see that the learning performance was almost the same even though we set $B=100$. 

\begin{figure}[tb]
    \begin{minipage}{1\hsize}
        \centering
        \includegraphics[width=5cm]{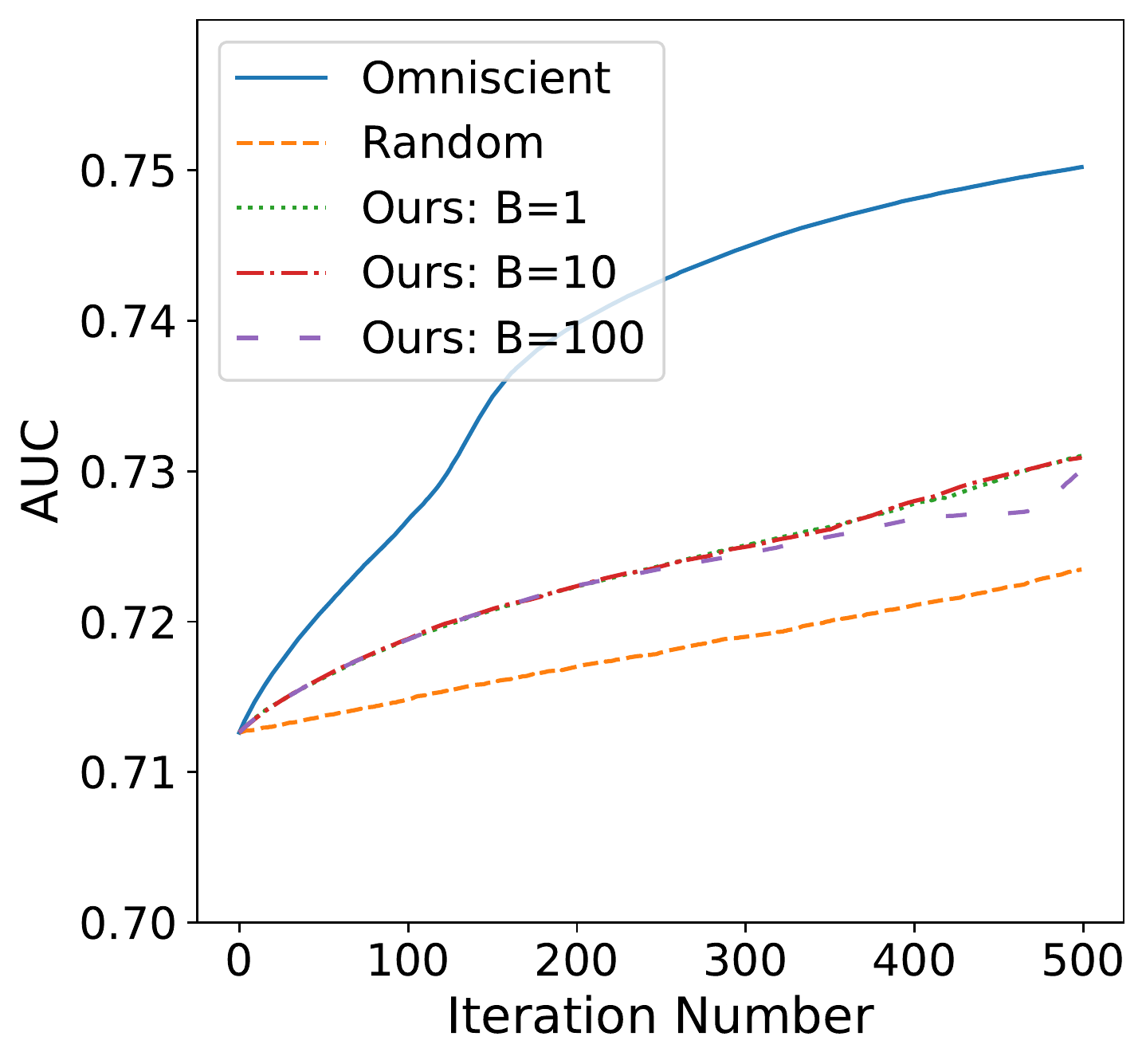}
        \subcaption{insect}
        \label{fig:insect_B_2}
    \end{minipage}
    \begin{minipage}{1\hsize}
        \centering
        \includegraphics[width=5cm]{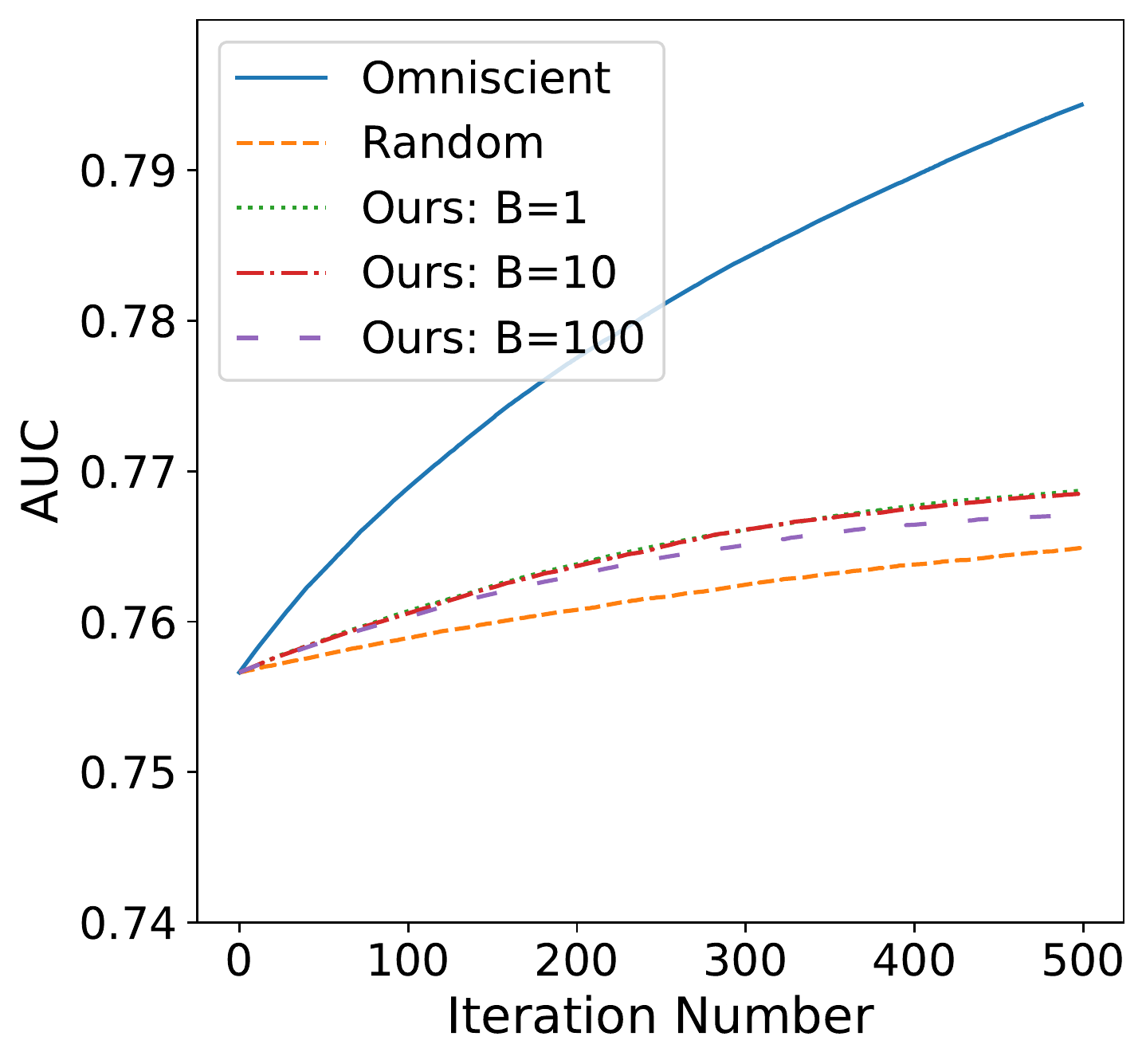}
        \subcaption{wine}
        \label{fig:wine_B_2}
    \end{minipage}
    \caption{Learning performance with different exam intervals ($B$)}
    \label{fig:insect_B}
\end{figure}

\subsubsection{Influence of exam size for model estimation.}

    We set $T$, the number of examples in an exam, to $100$ in the experiments. 
    If we use larger $T$, the model estimation accuracy can become high. 
    We evaluated the accuracy of the model estimation with different $T$. 
    \Zu{RMSE} shows the root mean square error (RMSE) between the true model $\bm{w}^\star$ and the estimated true model $\hat{\bm{w}}^\star$, and between the student models $\bm{W}$ and the estimated student models $\hat{\bm{W}}$. 
    We fixed the exam interval $B=1$ and we obtained the similar results for $B=10$ and $B=100$. 
    It is clear that the RMSE scores become smaller with increasing $T$.
    We also confirm that the estimation accuracy improves as the the iterations progress.

    \begin{figure}[tb]
            \begin{minipage}{1\hsize}
                \centering
                \includegraphics[width=5cm]{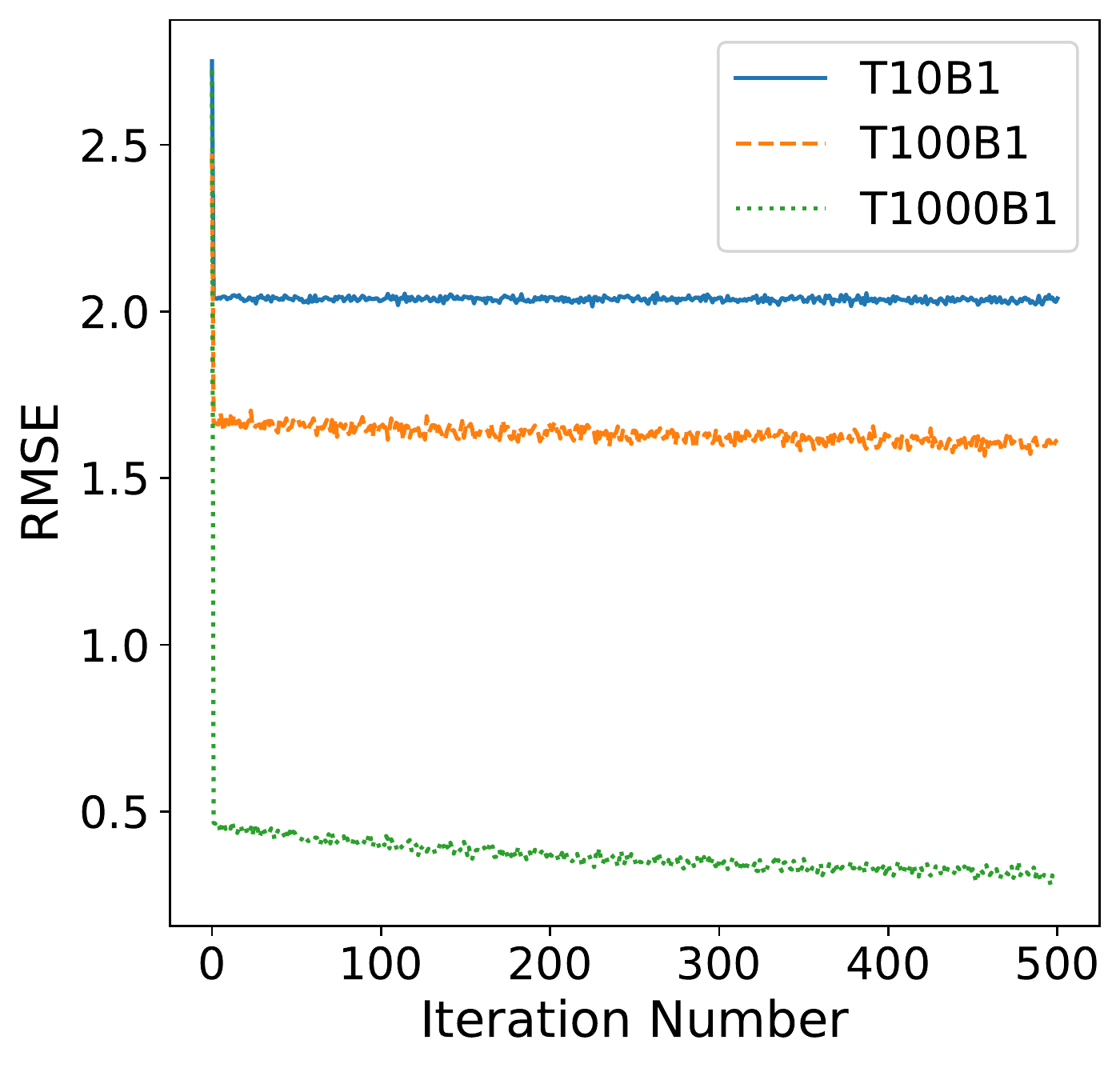}
                \subcaption{Estimated true model ($\hat{\bm{w}}^\star$)}
                \label{fig:rmse_w_star}
            \end{minipage}
            \begin{minipage}{1\hsize}
                \centering
                \includegraphics[width=5cm]{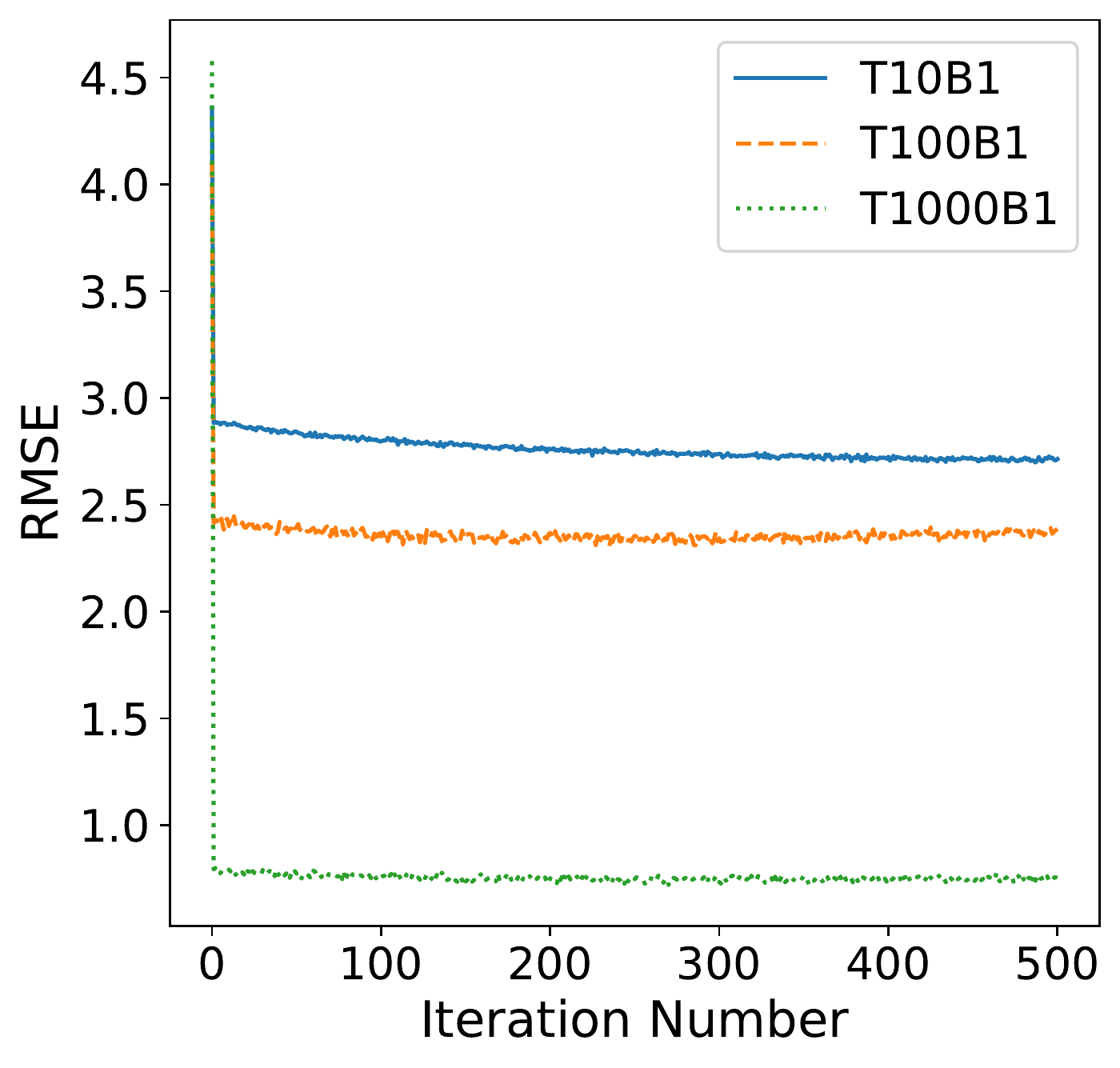}
                \subcaption{Estimated student models ($\hat{\bm{W}}^\star$)}
                \label{fig:rmse_W_star}
            \end{minipage}
            \caption{RMSE of estimated models with different sizes of teaching set ($T$)}
            \label{fig:RMSE}
    \end{figure}
    
\begin{figure*}[tb]
      \begin{minipage}{1\hsize}
        \centering
        \includegraphics[width=17cm]{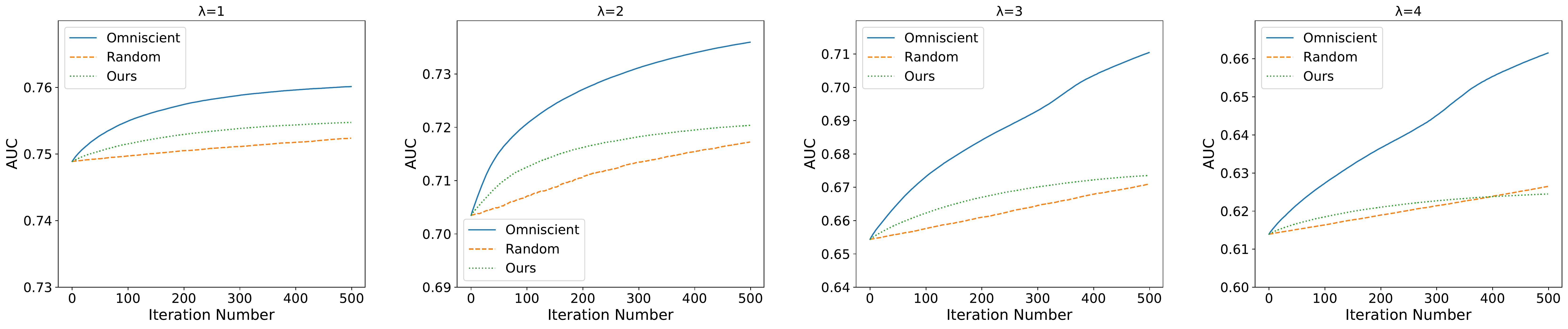}
        \subcaption{All students}
        \label{fig:insect_all}
      \end{minipage} \\
      \begin{minipage}{1\hsize}
        \centering
        \includegraphics[width=17cm]{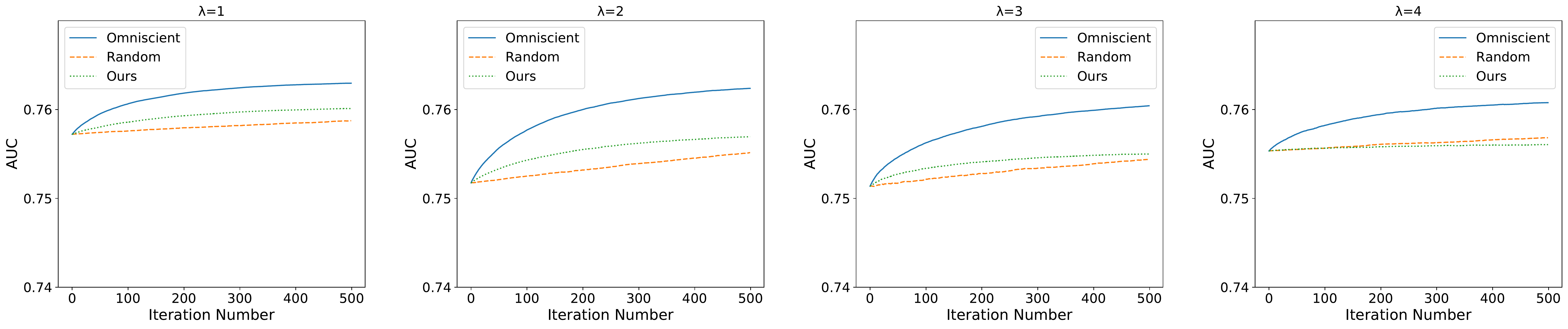}
        \subcaption{Ligh-level students}
        \label{fig:insect_high}
      \end{minipage} \\
      \begin{minipage}{1\hsize}
        \centering
        \includegraphics[width=17cm]{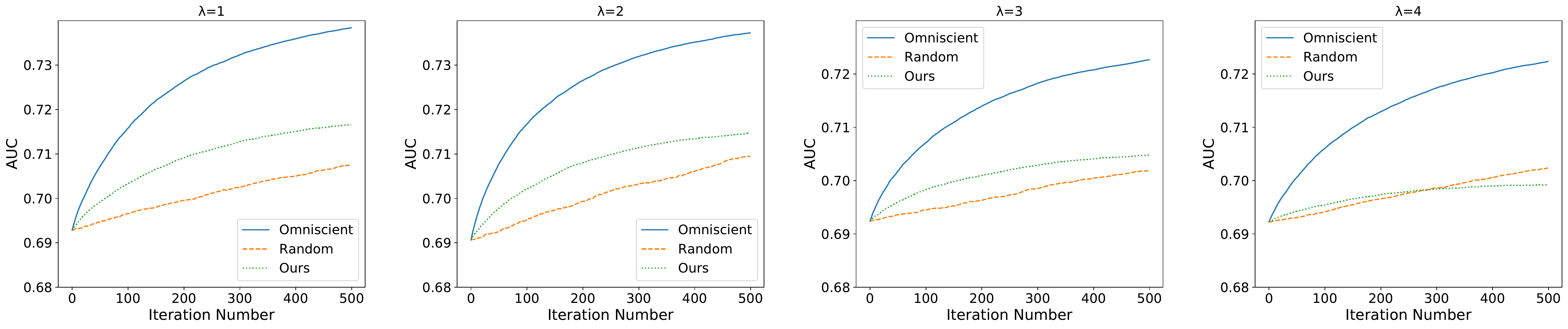}
        \subcaption{Middle-level students}
        \label{fig:insect_middle}
      \end{minipage}
      \begin{minipage}{1\hsize}
        \centering
        \includegraphics[width=17cm]{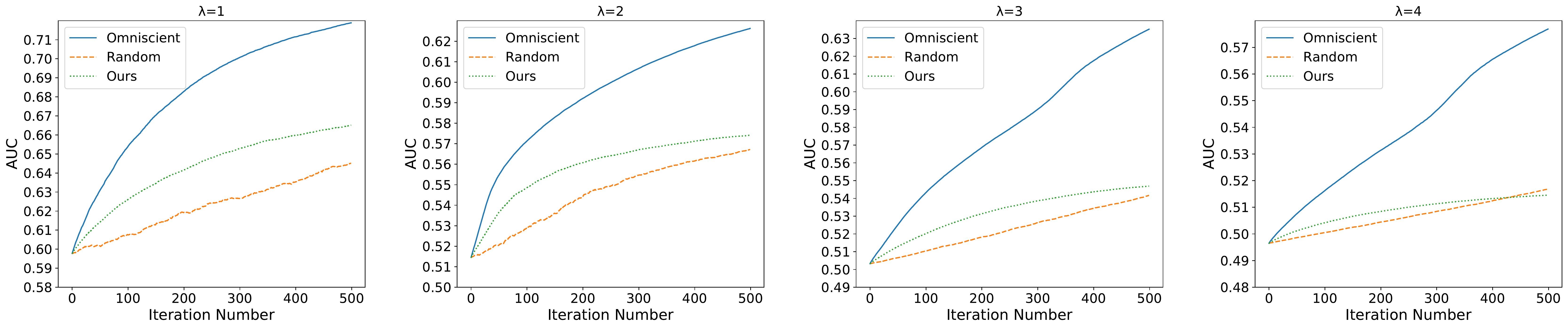}
        \subcaption{Low-level students}
        \label{fig:insect_low}
      \end{minipage}
      \caption{Learning performance with teaching algorithms (insect dataset)}
      \label{fig:insect}
  \end{figure*}

\begin{figure*}[tb]
      \begin{minipage}{1\hsize}
        \centering
        \includegraphics[width=17cm]{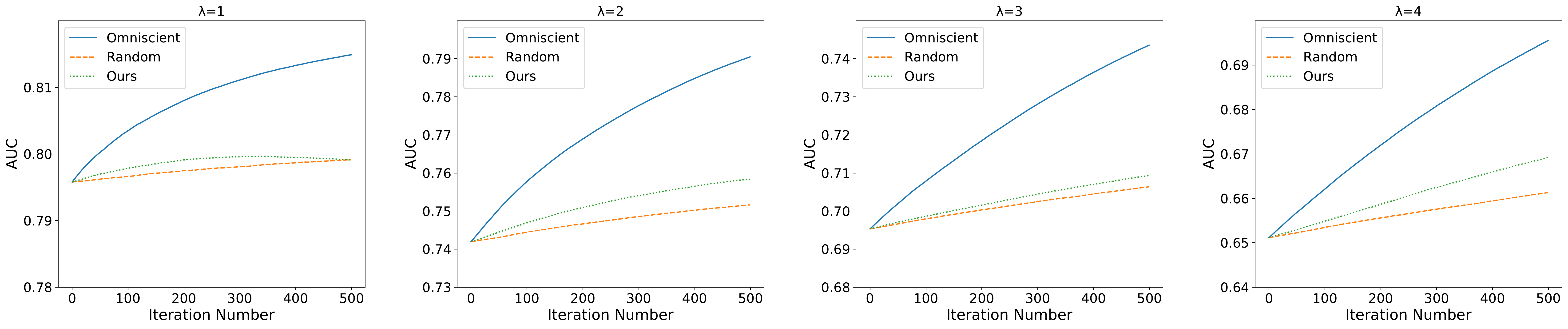}
        \subcaption{All students}
        \label{fig:wine_all}
      \end{minipage} \\
      \begin{minipage}{1\hsize}
        \centering
        \includegraphics[width=17cm]{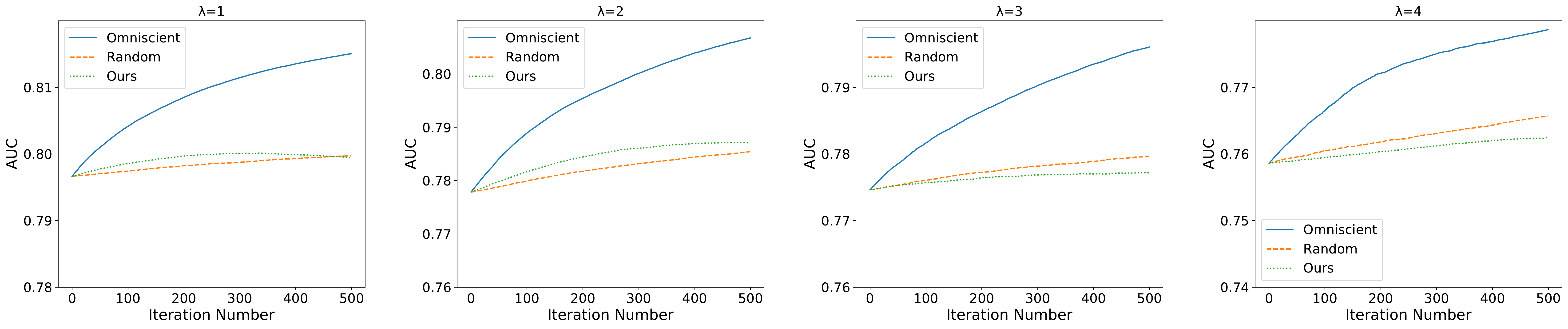}
        \subcaption{High-level students}
        \label{fig:wine_high}
      \end{minipage} \\
      \begin{minipage}{1\hsize}
        \centering
        \includegraphics[width=17cm]{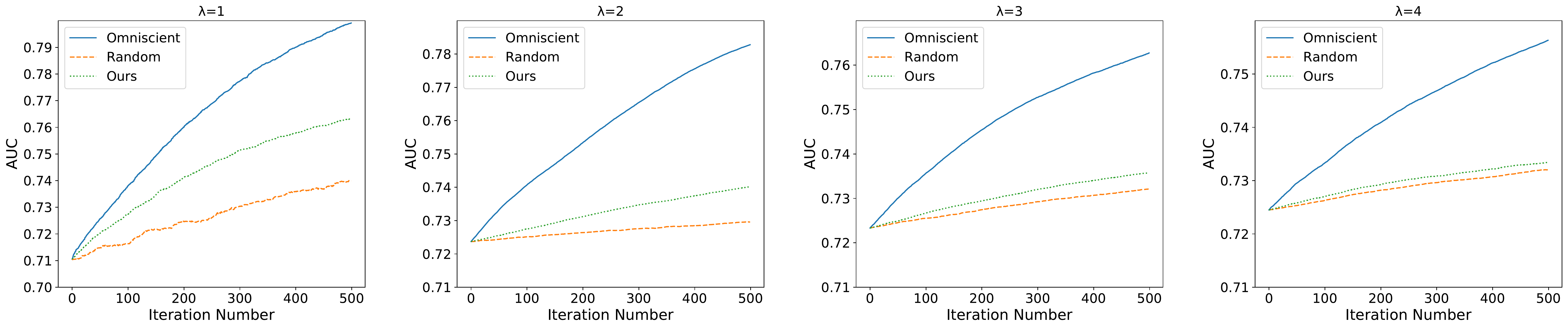}
        \subcaption{Middle-level students}
        \label{fig:wine_middle}
      \end{minipage}
      \begin{minipage}{1\hsize}
        \centering
        \includegraphics[width=17cm]{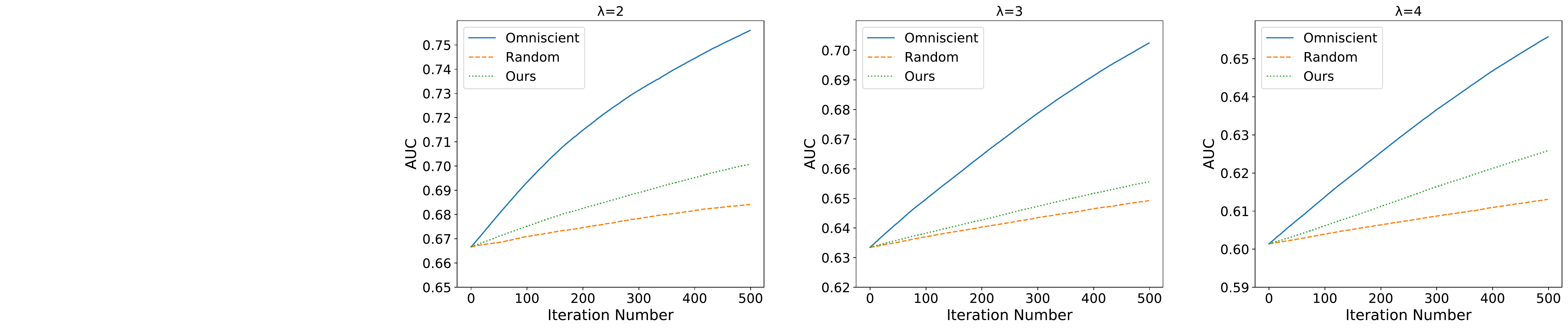}
        \subcaption{Low-level students}
        \label{fig:wine_low}
      \end{minipage}
      \caption{Learning performance with teaching algorithms (wine dataset); plot for $\lambda=1$ is not presented as there is no low-level students in this case.}
      \label{fig:wine}
  \end{figure*}

\section{Conclusion}
In this paper, we proposed a method for iterative machine teaching with an unlabeled teaching set. 
The results of the experiments showed that the proposed method achieved measurable learning efficiency even though it did not use true labels, and it was particularly effective for low-level students.
We also examined the effects of the size of an exam set ($T$) and the exam interval ($B$), and found that the number of student answers can be reduced by increasing $B$.

When performing experiments with humans, it may be possible to find human-specific properties (such as memory) that cannot be clarified by simulations. 
Considering such factors is a possible direction future studies.
In addition, the proposed method assumes that students and teachers share the same feature space; addressing the different feature space cases would be another area of focus for future work.

\footnotesize
\section*{Acknowledgements}

This work was supported by JSPS KAKENHI Grant Number JP18K18105 and JST PRESTO Grant Number JPMJPR19J9, Japan. 

\normalsize

\end{document}